\documentclass[11pt,a4paper]{article}
\usepackage[hyperref]{acl2017}
\usepackage{times}
\usepackage{latexsym}
\usepackage{slashbox}
\usepackage{graphicx}
\usepackage{url}
\usepackage{amssymb}
\usepackage{wasysym}

\aclfinalcopy % Uncomment this line for the final submission
\title{Analyzing Neural MT Search and Model Performance}

\author{Jan Niehues, Eunah Cho, Thanh-Le Ha, Alex Waibel \\
Institute for Anthropomatics and Robotics\\
Karlsruhe Institute of Technology, Germany \\ 
{\tt \{jan.niehues,eunah.cho,thanh-le.ha,alex.waibel\}@kit.edu} 
}

\date{}

\begin{document}
\maketitle
\begin{abstract}
In this paper, we offer an in-depth analysis about the modeling and search performance. We address the question if a more complex search algorithm is necessary. Furthermore, we investigate the question if more complex models which might only be applicable during rescoring are promising.

By separating the search space and the modeling using $n$-best list reranking, we analyze the influence of both parts of an NMT system independently. By comparing differently performing NMT systems, we show that the better translation is already in the search space of the translation systems with less performance. This results indicate that the current search algorithms are sufficient for the NMT systems. Furthermore, we could show that even a relatively small $n$-best list of $50$ hypotheses already contain notably better translations.

\end{abstract}

\section{Introduction}

Recent advances in NMT systems \cite{Bahdanau2014,cho2014learning} have shown impressive results in improving machine translation tasks. 
Not only it performed greatly in
recent machine translation campaigns~\cite{cettolo2016iwslt,bojar2016findings} measured in BLEU \cite{papineni2002bleu}, it is considered to be able to generate sentences with better fluency. 

Despite the successful results in translation performance, however, the optimality of the search algorithm in NMT has been left under-explored. 
In this work, we analyze the influence of search and modeling of an NMT system by evaluating them separately. 
We aim to demonstrate whether further research on the model development is more promising or the one on the search algorithm would be more beneficial. 

We attempt to simulate this by $n$-best rescoring using different models. For this, $n$-best lists are rescored by different models including the one which generated them. 
Additionally we build a configuration with all $n$-best lists joined, in order to see whether rescoring this joined $n$-best list using the same model would bring a performance boost.

\section{Related Work}
\label{refer} 
%Traditionally, 
There has been a number of works devoted to combine different systems from the same or different machine translation (MT) paradigms using $n$-best lists of hypotheses \cite{matusov2006computing,heafield2009machine,macherey2007empirical}. The hypotheses are aligned, combined and scored by a model to produce the best candidate according to a metric. 
There was a thorough analysis on how the size of $n$, the diversity of the outputs from different systems and performance of individual systems can affect the final translation of the system combination. \citet{hildebrand2008combination} examine the feature impact and the $n$-best list size of such a combination of phrase-based, hierarchical and example-based systems. \citet{gimpel2013systematic} show how diversity of the outputs and the size of the $n$-best lists determine the performance of the combined system.  
 
\citet{costa2007analysis} analyze the impact of the beam size used in statistical machine translation (SMT) systems.  
\citet{wisniewski2013oracle} conduct an in-depth analysis over several types of errors. Based on their proposal to effectively calculate oracle BLEU score for an SMT system, they can separate the errors due to the restriction of the search space (search error) from the errors due to models not good enough to cover the best translation (model error). 
Although this work is the closest to our work in terms of analysis methods, our work differs from theirs by addressing the issue focused on the NMT systems.

In \citet{neubig2015neural}, the size of the $n$-best list produced by a phrase-based SMT and rescored by an NMT is taken into account for an error investigation. 
The work also shows which types of errors from the phrase-based system can be corrected or improved after NMT rescoring. 
To the best of our knowledge, our work is the first to examine the impact of search and model performance in pure NMT systems. 
  
\subsection{Neural Machine Translation}
\label{nmt} 
Neural machine translation, whilst considered to be in the same direction with phrase-based SMT from a statistical perspective,
is actually separable from traditional SMT in terms of how it models the representation of source and target sentences as well as the translation relationship between them. 
In this section, we describe the general architecture of a NMT system in order to understand the needs and importance of such an analysis.
The NMT architecture described here is similar to the attention-based NMT from \citet{Bahdanau2014}.

An attentional NMT system consists of an encoder representing a source sentence and an attention-aware decoder that produces the translated sentence. 

The encoder which is comprised of bidirectional recurrent layers reads words from the source sentence and encodes them into annotation vectors. Each annotation vector contains the information of the source sentence related to the corresponding word from both forward and backward directions. 

A single layer featuring attention mechanism allows the decoder to decide which source words should take part in the prediction process of the current target word. Basically, attention layer examines a context vector of the source sentence which is weighted sum of all annotation vectors and normalized, where the weights reflect some relevance between previous target words and all the source words. 

The decoder, which is also recurrent-based, recursively generates the target candidates with their probabilities to be selected based on the context vector from the attention layer, the previous recurrent state and the embedding of the previously chosen word. 

The whole network is then trained in an end-to-end fashion to learn parameters which maximizes the likelihood between the outputs and the references. In the testing phase, a beam search is utilized to find the most probable target sequences giving the $n$-best list from the architecture.

We could see that in NMT, therefore, the model (e.g. the ways the encoder representing a source sentence or the attentional layer modeling attention mechanism)
and the search algorithm are one of the most important aspects to be analyzed.

\section{Search and Model Performance}
\label{anal} 
In this analysis we evaluate the search and modeling performance of NMT. In order to evaluate them individually, we need to separate the modeling errors and the search errors of the system. While the search in phrase-based MT was relatively complex, the search algorithm in NMT is relatively straightforward. In state-of-the-art system, a beam search algorithm is used with a small beam between $10-50$.

The goal of this work is to establish whether improvements on the NMT model itself is more promising or the ones on the search algorithm. If there are many search errors due to the pruning during decoding, a better search algorithm would be promising. In contrast, if there are relatively few search errors, further research on the model is more promising.

\subsection{Analysis Setup}

\begin{figure*}
\centering
\caption{\label{search}Search space analysis}
\includegraphics[width=0.6\textwidth]{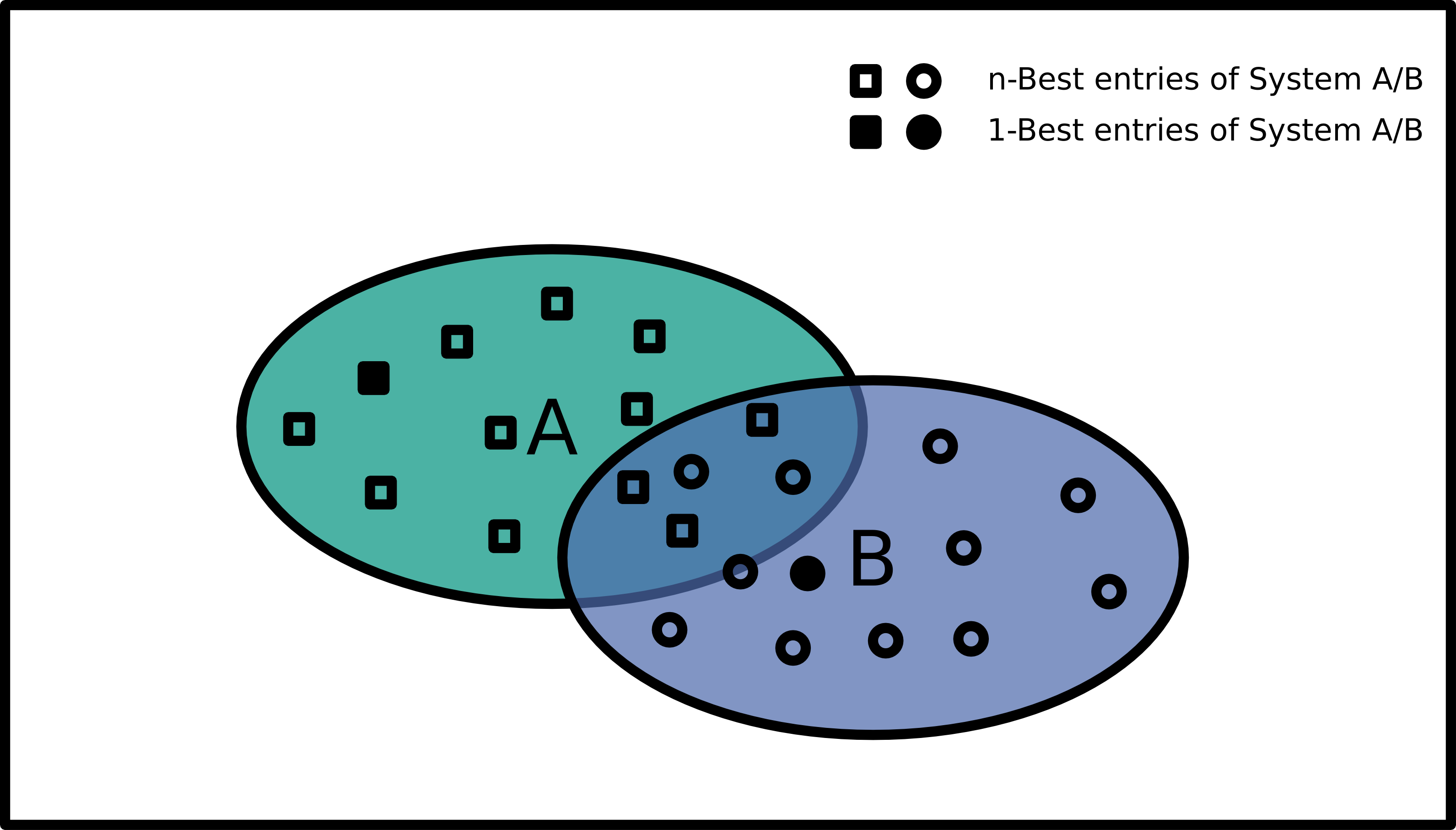}
\end{figure*}

A straightforward way would be to evaluate all possible hypotheses. In this case we do not have any search error and can directly measure the modeling errors. 
However this cannot be performed efficiently since the number of all possible hypotheses is very large. 
Therefore, we analyzed the performance of two or several systems with different performances.

In the experiments, for example, we have systems $A$ and $B$ where the translation performance of $A$ is better than the one of $B$. 
Then we approximated the search space of $A$ and $B$ by their $n$-best lists and evaluated the performance of each system in the search space of $A$ by scoring the $n$-best list with the model and selecting the hypothesis with the highest probability.
Figure \ref{search} shows the search spaces of system $A$ and $B$, approximated by their $n$-best hypotheses. Their 1-best entries are also marked accordingly. 

The question we address is why the system $B$ did select its best hypothesis $\CIRCLE$ and did not select the better-performant hypothesis $\blacksquare$. One reason might be that $\blacksquare$ is not in the system B's search space and therefore the system could not find it. The other reason might be that system $B$ prefers $\CIRCLE$ over $\blacksquare$. In this case, we need to improve the modelling.

If the performance of model $B$ on the $n$-best list of $A$ is better than the initial score of $B$, it suggests that the model $B$ is able to select a better hypothesis and therefore the search is not optimal. On the other hand, if the performance is similar, it means that $B$ is not able to select a better hypothesis, even though there are better ones according to the evaluation metric. 

In the experiments, we used two different ways of constructing the models $A$ and $B$. In a first series of experiments, we used the single best system as well as ensemble systems. In a second series, we used systems using different ways to generate the translations. Details of the systems will be given in the Section \ref{sysdescript}. 

\section{System Description}
\label{sysdescript} 
Our German$\leftrightarrow$English NMT systems are built using an encoder-decoder framework with attention mechanism, \texttt{nematus}.\footnote{\small \tt{https://github.com/rsennrich/nematus}}
Byte pair encoding (BPE) is used in order to generate sub-word units \cite{sennrich2015neural}. 
Long sentences whose sentence length exceeds 50 words are exempted from the training. 
We use minibatch size 80 and sentences are shuffled within every minibatch. 
Word embedding of size 500 is applied, with hidden layers of size 1024. 
Dropout is applied at every layer with the probability 0.2 in the embedding and hidden layers and 0.1 in the input and output layers. 
Our models are trained with Adadelta \cite{zeiler2012adadelta} and the gradient norm is clipped to 1.0. 
For the single models, we apply the early stopping based on the validation score.

The baseline system is trained on the WMT parallel data, namely EPPS, NC, CommonCrawl and TED corpus. 
As validation data we used the newstest13 set from IWSLT evaluation campaign. Therefore, this data is from TED talks.  
Test is applied on two domains. First domain is TED talks, same as the optimization set. We use newstest14 for this testing.  
Another domain is telephone conversation and we used MSLT \cite{federmann2016microsoft} for testing. 
Since no exact genre-matching development data is published for the evaluation campaign \cite{cettolo2016iwslt}, 
we used the TED-optimized system for the MSLT testing.  
For each experiment, we also offer oracle BLEU scores on the $n$-best lists, calculated using multeval \cite{clark2011}.

\subsection{Configurations} 

We tried different system configurations to generate and rescore the $n$-best lists. 
By using 40K operations of BPE we had \textit{SmallVoc} configuration, and with 80K \textit{BigVoc} configuration. 
In \textit{SmallVoc.rev}, target sentence are generated in the reversed order. In \textit{SmallVoc.mix}, target side corpus is joined with the source side corpus to form a mixed input as described in \citet{cho2016adaptation}. 
We build an NMT system which takes pre-translation from a PBMT system following the work in \citet{Niehues2016}, which will be referred as \textit{PrePBMT}. 
A configuration using more monolingual data for this training is called \textit{PrePBMT.large}. 
In \textit{Union}, we use the joined $n$-best lists from different systems. 

\subsection{$n$-best list} 

All $n$-best lists are generated for $n = 50$ from a standard beam search. 
The size of $n$-best lists are limited due to time and computational limitation. In our preliminary experiments where we increased $n$-best list from 1 to 50, 
it did not significantly change the performance of one model. Therefore, in this work, we approximate the $50$-best lists our search space and conducted the analysis. 
By doing so, we also aim to give a practical analysis on a model vs. search performance comparison in NMT and useful guidelines from it.

\section{Analysis on the Results}
\label{resul} 
In this section, we discuss the experimental results and detailed analysis.  
In the first part, we discuss the results of the experiments on the baseline systems. In the second part, we combine NMT systems that use different text representations. 

\subsection{NMT Baseline Systems}
In this section, we analyze the performance of baseline systems. It largely breaks down to two tasks: TED and MSLT translation. 

\subsubsection{TED translation} 
Table \ref{de-en_ted_single} shows the baseline system performance on the TED translation task, from German to English. 
The table is showing translation performance of reranking each $n$-best list using different models.

 \begin{table}[htb]
\centering
   \begin{tabular}{l|cccccc} \hline
   
  \backslashbox{$n$-best}{Model} & Single & Ensemble & Oracle \\
 \hline
Single & 31.96 & 32.37 & 41.81\\
Ensemble & 32.09 & 32.41 & 42.31\\ 
Union & 31.95 & 32.39 & 44.55\\
 \hline
   \end{tabular}
 \caption{\label{de-en_ted_single} Baseline: TED German$\rightarrow$English}
 \end{table}

 For the \textit{Single} system, we took the best-performant \textit{BigVoc} system. The \textit{Ensemble} system is then generated by combining several training steps of the single system training. The \textit{Union} $n$-best list is the joined $n$-best list of all the individual systems used in the ensemble. 
For building a \textit{Ensemble} system, we combine different models from several time steps of \textit{Single} training. Then in the softmax operation, normalized probabilities of each word are considered. 
As mentioned earlier, we also offer the oracle BLEU scores given each $n$-best list. 
 
 \paragraph{Model performance} 
 As shown in the table, we can improve the translation performance by 0.5 BLEU point by using the \textit{Ensemble} system to rescore $n$-best list generated by the same system, compared to the same case for \textit{Single} system. The main contribution for this improvement seems to be the better modeling. 
 When we use the \textit{Single} model to rescore the \textit{Ensemble} or \textit{Union} $n$-best list, we get mainly the same performance. 
 Thus, the reason for the relatively lower performance of the \textit{Single} system is considered to be that it does not model the translation probabilities better, not because it does not find better translations. The oracle scores indicate the similar trend. 
When the $n$-best list is large (\textit{Union} setup), we have better translations in the $n$-best list. However, these hypotheses were not selected by either of the models. 

 \paragraph{Search performance} 
The numbers in Table \ref{de-en_ted_single} suggests that the search is well-performant in NMT. 
For example, when we use the \textit{Ensemble} model to rescore the \textit{Single} $n$-best list, the translation performance reaches 32.37 BLEU points. 
At the same time, when we use the same model to rescore the \textit{Ensemble} $n$-best list, we achieve a similar performance. 

\subsubsection{MSLT translation} 

The performance on the single system on the MSLT task \cite{federmann2016microsoft} is shown in Table \ref{de-en_mslt_single}.

 \begin{table}[htb]
\centering
   \begin{tabular}{l|cccccc} \hline 
 \backslashbox{$n$-best}{Model} & Single & Ensemble & Oracle \\
 \hline
Single & 34.63 & 38.35 & 53.85\\
Ensemble & 35.94 & 38.80 & 56.46\\ 
 \hline 
   \end{tabular}
 \caption{\label{de-en_mslt_single} Baseline: MSLT German$\rightarrow$English}
 \end{table}

In this task, rescoring \textit{Single} $n$-best list using the same model itself performs around 4 BLEU worse than rescoring \textit{Ensemble} $n$-best list using the \textit{Ensemble} model. 
Also, we can observe that the \textit{Single} model performs better when using the $n$-best list of the \textit{Ensemble} model. 

We find two explanations for this improvement. 
A) The \textit{Ensemble} $n$-best list contains better-performing hypotheses that the \textit{Single} model did not find during the search. 
Or alternatively, B) 
the \textit{Ensemble} $n$-best list does not contain the hypotheses that are good according to the \textit{Single} model but not according to the evaluation metric. 
In this case, the model would select different hypotheses. 

In order to locate search error, we evaluated and compared the model score of hypothesis chosen from different $n$-best lists. 
Only in 2.5\% of the chosen hypotheses, the score of the hypothesis selected from the \textit{Ensemble} $n$-best list is higher than the one from the \textit{Single} $n$-best list. 
Thus, we have a search error only in these cases. 

In contrast, in 90.7\% of the sentences, the score from the \textit{Single} $n$-best list is higher. 
The main reason for the improvement, therefore, is not considered to be better search. 
Rather, the search space by the \textit{Ensemble} system does not contain the worse-performing translations which are highly ranked by the \textit{Single} system.

The $n$-best lists of the \textit{Single} model contains well-performing translations. 
For example, the performance achieved when using the \textit{Ensemble} model to rescore the \textit{Single} $n$-best list 
is almost similar to the one achieved when applying the same model on the \textit{Ensemble} $n$-best list. 
This performance is nearly 4 BLEU points better than rescoring the same $n$-best list using the \textit{Single} model. 

While the performance of the \textit{Ensemble} model on both $n$-best lists is similar, interestingly, the oracle score of the \textit{Ensemble} $n$-best list is clearly higher. 
Therefore, the models seem not able to select better translations in the \textit{Ensemble} $n$-best list compared to the \textit{Single} $n$-best list.

\subsection{NMT Text Representation Systems}

As a next line of experiment, we combine NMT systems that use different text representations. 

\subsubsection{TED translation} 

Table \ref{de-en_TED_Multi} lists the systems used in the experiment and their performance on the TED task. 

 \begin{table*}[htb]
\centering
   \begin{tabular}{l|cccccc} \hline 
  \backslashbox{$n$-best list}{Model} & SmallVoc & SmallVoc.rev & BigVoc & PrePBMT  & All & Oracle\\
 \hline
SmallVoc  & \textbf{31.74} & 32.17 & 32.62 & 32.55 & 33.03 & 41.82\\
SmallVoc.rev & 32.24 & \textbf{31.28} & 32.58 & 32.06 & 32.93 & 40.97\\
BigVoc  & 32.57 & 32.50 & \textbf{32.41} & 32.63 & 33.26 & 42.31\\
PrePBMT & 32.19  & 31.97 & 32.53 & \textbf{31.41} & 32.65 & 40.67\\
Union & \textbf{31.83} & \textbf{31.27} & \textbf{32.42} & \textbf{31.39} &  33.24 & 46.58\\
 \hline 
   \end{tabular}
 \caption{\label{de-en_TED_Multi} Text representation systems: TED German$\rightarrow$English}
 \end{table*}

We can observe that the results of \textit{Union} rescored by each model is similar to the performance of the model's $n$-best list rescoring, as marked in bold letters in each column. 
Considering that the \textit{Union} $n$-best list is considerably larger, it seems again that the model can find the best hypothesis according to the model. 

In contrast, if we use all models (\textit{All}) by using sum of log probabilities of all models to rescore the $n$-best lists, we achieve similar performance for all $n$-best lists. 
Thus, it seems that all $50$-best lists contain already very good hypotheses. 
Only the $n$-best list of the \textit{PrePBMT} system seems to contain relatively worse options. 
This is also shown by the oracle scores. 
One reason could be that the pre-translation by the PBMT system is guiding the search and therefore the $n$-best list contains relatively limited variety.

In addition, we observe that the performance of each model on its own $n$-best list is considerably worse than the model rescoring other $n$-best lists. 
This can be explained by the following phenomena: 
some translations of a system $A$ are highly-ranked by the model itself, but not by the others. 
Therefore, they are selected by the system $A$ but not in the $n$-best lists of the other systems. If they are in the $n$-best list, e.g. in the $n$-best lists of the system $A$ and in the \textit{Union}, they will be selected only when using the system $A$, leading to worse performance in BLEU. 
In contrast, if we use different $n$-best lists, the translation performance is better. 
 
\paragraph{English$\rightarrow$German}

In addition, we extend this experiment to another language direction. 
Table \ref{en-de_MSLT_Multi} shows the results when the same experiment is applied to En-De TED task. 

   \begin{table*}[htb]
\centering
   \begin{tabular}{l|cccccc} \hline  
  \backslashbox{$n$-best list}{Model} & SmallVoc.mix & BigVoc & PrePBMT & PrePBMT.large  & All & Oracle\\
 \hline
 SmallVoc.mix & \textbf{26.19} & 27.09 & 26.93 & 27.03 & 27.12 & 33.71\\
 BigVoc & 26.97 & \textbf{27.28} & 27.26 & 27.12 & 27.48 & 34.16\\
 PrePBMT & 26.96 & 27.00 & \textbf{26.44} & 27.15 & 27.14 & 32.95\\
 PrePBMT.large & 27.25 & 27.47 & 26.85 & \textbf{27.03} & 27.41& 33.78\\
Union & \textbf{26.25} & \textbf{27.28} &  \textbf{26.44}  & \textbf{27.03} &27.76  & 38.95\\
 \hline 
   \end{tabular}
 \caption{\label{en-de_MSLT_Multi} Text representation systems: TED English$\rightarrow$German}
 \end{table*}

Here the same phenomena is observed. 
Again, the \textit{Union} $n$-best list does not improve the translation quality. 
Nonetheless, the oracle score is significantly higher indicating that the model finds the better hypotheses. 
Furthermore, the $n$-best lists already contain better hypotheses which can be chosen using better models, i.g. the combination of all models.

\subsubsection{MSLT translation} 

 Table \ref{de-en_MSLT_Multi} shows the similar results when the same experiments are applied to the MSLT task. 
The \textit{Union} configuration performs similar to rescoring using the same model, while performing considerably worse than the case where the same $n$-best list rescored by other models.

   \begin{table*}[htb]
\centering
   \begin{tabular}{l|cccccc} \hline  
  \backslashbox{$n$-best list}{Model} & SmallVoc & SmallVoc.rev & BigVoc & PrePBMT  & All & Oracle\\
 \hline
SmallVoc  & \textbf{37.90} & 39.50 & 39.35 & 39.05 & 40.30 & 56.41\\
SmallVoc.rev & 39.74 & \textbf{38.72} & 39.92 & 39.94 & 40.80 & 56.82\\
BigVoc  & 38.73 & 39.61 & \textbf{38.80} & 39.51 & 40.25 & 56.46 \\
PrePBMT & 38.91 & 39.68 & 39.36 & \textbf{38.33} & 40.24 & 54.44\\
Union & \textbf{37.92} & \textbf{38.65} & \textbf{38.81} & \textbf{38.33} & 40.66 & 63.09\\
 \hline 
   \end{tabular}
 \caption{\label{de-en_MSLT_Multi} Text representation systems: MSLT German$\rightarrow$English}
 \end{table*}

\section{Conclusion}
\label{conclu} 

Our experiments on two language pairs and two different tasks showed that there are only few search errors in the state-of-the-art NMT systems. 
Even when better hypotheses are added in the $n$-best list, the models do not select a different hypothesis. Thus, the search algorithms seem to be sufficient. 
Furthermore, we showed that a relatively small $n$-best list of 50 entries already contains notably better translation hypotheses. This result indicates that improving rescoring models are promising for performance boost. 
In this work, we showed that it is often sufficient to use a model in rescoring only. 
This finding also motivates 
the development of models which are challenging to use directly during the decoding, such as bi-directional decoders.

\section*{Acknowledgments}

The project leading to this application has received funding from the European Union's Horizon 2020 research and innovation programme under grant agreement n$^\circ$ 645452.
The research by Thanh-Le Ha was supported by Ministry of Science, Research and the Arts Baden-W\"urttemberg.
This work was supported by the Carl-Zeiss-Stiftung.

\bibliographystyle{acl_natbib}
\bibliography{acl2017}

\end{document}